# PINN-EMFNet: PINN-based and Enhanced Multi-Scale Feature Fusion Network for Breast Ultrasound Images Segmentation

Jiajun Ding[a], Beiyao Zhu[c], Wenjie Wang[b], Shurong Zhang[d], Dian Zhu[a*], Zhao Liu[a*]

[a] *School of Design, Shanghai Jiao Tong University, Shanghai, PR China*

[b] *School of Mechanical Engineering, Shanghai Jiao Tong University, Shanghai, PR China*

[c] *Shanghai Ninth People's Hospital, Shanghai Jiao Tong University School of Medicine, Shanghai, P.R. China.*

[d] *Department of Sports Medicine, Huashan Hospital, Fudan University, P.R. China.*

## Abstract

With the rapid development of deep learning and computer vision technologies, medical image segmentation plays a crucial role in the early diagnosis of breast cancer. However, due to the characteristics of breast ultrasound images, such as low contrast, speckle noise, and the highly diverse morphology of tumors, existing segmentation methods exhibit significant limitations in terms of accuracy and robustness. To address these challenges, this study proposes a PINN-based and Enhanced Multi-Scale Feature Fusion Network. The network introduces a Hierarchical Aggregation Encoder in the backbone, which efficiently integrates and globally models multi-scale features through several structural innovations and a novel PCAM module. In the decoder section, a Multi-Scale Feature Refinement Decoder is employed, which, combined with a Multi-Scale Supervision Mechanism and a correction module, significantly improves segmentation accuracy and adaptability. Additionally, the loss function incorporating the PINN mechanism introduces physical constraints during the segmentation process, enhancing the model's ability to accurately delineate tumor boundaries. Comprehensive evaluations on two publicly available breast ultrasound datasets, BUSIS and BUSI, demonstrate that the proposed method outperforms previous segmentation approaches in terms of segmentation accuracy and robustness, particularly under conditions of complex noise and low contrast, effectively improving the accuracy and reliability of tumor segmentation. This method provides a more precise and robust solution for computer-aided diagnosis of breast ultrasound images.

\* The first two authors contributed equally. Corresponding authors: Zhao Liu: hotlz@sjtu.edu.cn and Dian Zhu: zhudian@sjtu.edu.cn.

# 1. Introduction

Breast cancer is one of the most common malignant tumors among women, with approximately 2.3 million women diagnosed annually worldwide, making it one of the leading causes of death among women[1]. Early diagnosis of breast cancer is crucial in reducing patient mortality. Ultrasound imaging, due to its safety, high sensitivity, real-time capabilities, and low cost, has been widely used for screening and diagnosing breast tumors[2]. In particular, compared to mammography, ultrasound imaging offers higher sensitivity for detecting dense breast tissue and does not pose any radiation risk[3].

Image segmentation is essential in ultrasound imaging for breast cancer diagnosis. It not only aids in identifying and quantifying the size and shape of tumors but also provides vital information for tumor grading and feature analysis[3, 4]. Accurate image segmentation can improve breast cancer detection rates, particularly at early stages, which is critical for enhancing patient prognosis and reducing mortality rates. Moreover, segmentation techniques assist clinicians in making more precise diagnoses and treatment plans, including determining the scope of surgical excision and the localization of radiotherapy[5].

Despite its significance in breast cancer diagnosis, manual segmentation faces several challenges. Firstly, the inherent noise and low contrast of ultrasound images result in blurred edges or even the loss of edge information, which increases the difficulty of segmentation. Secondly, the shape and size of tumors vary significantly among patients, requiring segmentation algorithms to possess a high degree of adaptability and flexibility[6]. Additionally, manual segmentation relies heavily on the operator's experience and skills, leading to subjectivity and inconsistencies in the results[7].

With the rapid development of deep learning and computer vision technologies, medical image analysis is undergoing transformative advancements[8, 9]. In medical image segmentation, computer vision technologies assist physicians in rapidly and accurately extracting lesion areas, thereby supporting early diagnosis and subsequent treatment[10, 11].

Currently, breast ultrasound image segmentation methods mainly include traditional techniques and deep learning-based approaches. Traditional segmentation methods, such as threshold-based segmentation, region growing methods, and Active Contour Models (ACM), can achieve effective segmentation under specific conditions but typically require significant manual adjustment and supervision by experts. They also perform poorly when dealing with complex noise and low-contrast images[12, 13]. These methods rely on regularized features and struggle to address the diverse shapes and sizes of breast tumors. Moreover, they are sensitive to weak boundaries and noise, resulting in unstable segmentation outcomes. In contrast, deep learning-based segmentation networks, particularly those based on Convolutional Neural Networks (CNNs), can automatically extract multi-level features, significantly improving segmentation accuracy and robustness[14, 15]. Furthermore, deep learning models possess excellent generalization capabilities, allowing them to adapt to variations in images from different devices and environments, thus reducing the impact of human interference. These advantages have made deep learning a frontier approach in the research of breast ultrasound image segmentation.

Figure 1 illustrates a breast ultrasound image, where the tumor region (target segmentation area) is marked with a green contour. It is evident that the tumor region in ultrasound images closely resembles other regions, making manual differentiation quite challenging. Thus, the precise segmentation of tumor regions in breast ultrasound images remains a highly challenging task. These challenges mainly arise from the low contrast of the images, the interference of speckle noise, and the high variability in tumor shape and size across different patients[16-18]. The noise in ultrasound images, especially speckle noise, not only makes it difficult to distinguish the boundaries between tumors and normal tissue, but it also leads to the extraction of irrelevant features during algorithm training, thus reducing segmentation accuracy.

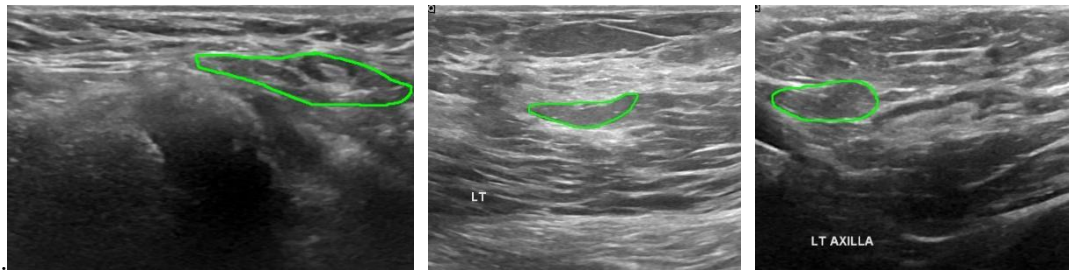

Figure 1. Breast Ultrasound Image

Deep learning-based automatic segmentation techniques have been widely applied to breast ultrasound image segmentation, with Convolutional Neural Networks and their derivative architectures, such as UNet, becoming one of the most popular segmentation methods due to their ability to automatically extract features[19, 20]. UNet and its improved versions (e.g., UNet++) demonstrate remarkable capabilities in medical image segmentation tasks, particularly in capturing the boundaries and shapes of breast tumors, by leveraging skip connections and multi-scale feature fusion[15, 21, 22]. Additionally, multi-task learning-based networks, such as JCS and ECSUNet, have been used for simultaneous classification and segmentation, enabling joint diagnosis[23]. The CSwin-PNet model proposes a pyramid network that combines CNNs with Swin Transformers to improve the segmentation performance of breast lesions in ultrasound images[24]. The MDF-Net model adopts a two-stage end-to-end architecture, with the backbone subnetwork extended from UNet++, featuring a simplified skip-path structure that connects features across adjacent scales[6]. The refinement subnetwork utilizes a structurally optimized MDF mechanism to integrate preliminary segmentation information at coarse scales and explore inter-subject variations at finer scales. The self-attention mechanism of transformer, which computes the relationships between various positions within an input feature sequence, allows the network to focus on long-range dependencies across different regions. This overcomes the limitations of CNNs in extracting distant dependencies[25]. As a result, Transformer-based networks have also been introduced into medical image segmentation. For instance, networks like TransUNet integrate the self-attention mechanism with convolutional networks, further enhancing feature representation capabilities. In addition to TransUNet[26], other Transformer-based networks, such as HiFormer-L[27], have also achieved promising results in medical image segmentation. The Swin-Transformer, which features a hierarchical design and a sliding window mechanism, not only enables global feature modeling but also significantly reduces computational complexity[28]. Furthermore, UNETR applies Transformers to the entire encoder section, further demonstrating the potential of Transformers in medical image segmentation[29].

Despite their advantages, these methods still face limitations when processing breast ultrasound images, particularly in terms of accurate tumor boundary localization and noise suppression capabilities[30]. Speckle noise, which is prevalent in ultrasound images, can propagate across different scales and convolution layers, significantly impacting segmentation accuracy. The convolution operations within CNNs predominantly concentrate on local image regions, making it challenging to capture global contextual information. When dealing with ultrasound images that have complex structures, CNNs may fail to fully understand the global features, which negatively affects segmentation precision[31]. Moreover, the variability in tumor shapes and sizes in ultrasound images further exacerbates this limitation, resulting in inaccurate segmentation outcomes[32]. While Transformers offer advantages in global feature modeling, their performance can also be compromised by high-frequency noise interference in breast ultrasound images, which reduces segmentation accuracy. Transformers tend to capture global context information, but in the case of ultrasound images, which are rich in fine details, they may fail to adequately capture local features. Additionally, the self-attention mechanism in Transformer models, when applied to high-resolution images, leads to quadratic growth in computational complexity, which significantly increases the demand for computational resources and storage. This poses a major limitation for their application in resource-constrained environments. In summary, existing breast ultrasound image segmentation methods face several limitations when addressing the challenges of ultrasound image segmentation. First, the accuracy of tumor boundary localization is insufficient, which negatively impacts segmentation performance. Second, the ability to suppress speckle noise is limited, leading to a decline in segmentation accuracy. Additionally, CNNs have limitations in capturing both global and local features, making it difficult to effectively handle the diversity of complex structures and tumor morphologies. While Transformers can model global information, their high computational complexity increases the demand for computational resources when processing high-resolution images, thus limiting their practical applicability.

To improve the accuracy and robustness of breast tumor segmentation in ultrasound images, this study proposes the PINN-based and Enhanced Multi-Scale Feature Fusion Network. The network framework mainly enhances segmentation accuracy by efficiently fusing global modeling, local detail capture, and multi-scale features, thereby reducing errors caused by noise and overcoming the limitations of existing methods in tumor boundary localization, noise suppression, and global and local feature extraction. Specifically, the framework consists of two main stages: the Encoder and the Decoder.

In the Encoder stage, a novel architecture called the Hierarchical Aggregation Encoder (HA-Encoder) is introduced. This backbone network focuses on the deep integration of features from different layers, which significantly improves tumor boundary localization, especially in cases with complex structures and diverse tumor morphologies. Through the effective fusion of multi-level features, the HA-Encoder provides more precise descriptions of tumor boundaries in ultrasound images, addressing the accuracy issues encountered in traditional methods for boundary recognition. Additionally, through structural innovations and the newly proposed Parallel Convolutional Attention Module (PCAM), the network achieves efficient fusion of multi-scale features and global modeling. The PCAM is better at capturing global information in images and enhancing local details, solving the problem of balancing global and local features in traditional Convolutional Neural Networks (CNNs). This balance between global and local features enables the model to more accurately handle tumor shape diversity and complex structures, thus improving segmentation performance. The Multi-Scale Feature Refinement Decoder (MSFR-Decoder) in the Decoder stage processes the multi-scale features extracted by the Encoder. The decoder first integrates the features generated at different scales by the backbone network through a multi-scale supervision mechanism (MSM), further enhancing the model's sensitivity to multi-scale information. This helps effectively overcome noise interference and improves segmentation accuracy. Next, in the refinement module, features obtained through the Sigmoid activation function are further fused across different scales, allowing the model to adaptively improve segmentation quality, especially in regions with blurred boundaries and strong noise. Finally, in the loss function section, to enhance the physical consistency of the model, this study incorporates the Physics-Informed Neural Network (PINN) mechanism[33, 34], which further improves segmentation accuracy and consistency. By introducing the PINN mechanism, the model can consider physical constraints during training, enhancing its ability to correctly segment the target area, especially in ultrasound images where tumor morphology varies significantly and noise is prevalent. The PINN mechanism effectively reduces segmentation errors, ensuring the stability and consistency of the segmentation results.

In conclusion, the PINN-based and Enhanced Multi-Scale Feature Fusion Network proposed in this study effectively addresses several issues in current breast ultrasound image segmentation, including inaccurate tumor boundary localization, noise interference, insufficient global and local feature extraction, and high computational complexity. This approach significantly enhances segmentation accuracy and robustness.

Through these innovations, the proposed network significantly enhances segmentation performance in ultrasound images by improving multi-scale feature extraction, noise suppression, and boundary refinement. The network was comprehensively evaluated on two publicly available breast ultrasound lesion datasets, BUSIS[35] and BUSI[36]. Specifically, on BUSIS, the network achieves an average Dice Similarity Coefficient of $84.82 \pm 3.33$, an Intersection over Union of $76.83 \pm 4.14$, Sensitivity of $85.15 \pm 3.40$, and Specificity of $99.49 \pm 0.22$. On the BUSI dataset, it attains a DSC of $78.28 \pm 5.13$, IoU of $70.29 \pm 5.10$, Sensitivity of $80.54 \pm 4.45$, and Specificity of $98.17 \pm 1.18$. Experimental results show that, compared to previous methods, our approach achieves significant improvements in segmentation accuracy, robustness.

Our contributions can be summarized as follows:
- This study proposes a backbone network architecture called the Hierarchical Aggregation Encoder. This encoder enables global modeling while efficiently capturing local details and integrating multi-scale features, offering significant advantages in multi-scale feature perception and fusion.
- The Multi-Scale Feature Refinement Decoder is introduced into the proposed method, significantly enhancing segmentation accuracy through a refined multi-scale supervision mechanism.
- In the loss function, this study incorporates the Physics-Informed Neural Network, which applies physical constraints to the segmentation results. This mechanism helps the network maintain physical consistency when learning tumor boundaries, improving the accuracy and consistency of segmentation in the target regions.
- Experimental results demonstrate that the proposed PINN-EMFNet method outperforms current state-of-the-art methods in segmentation on public datasets.

The organization of this study is as follows: Section 2 introduces the proposed PINN-EMFNet; Section 3 presents the datasets, evaluation metrics, and experimental results; Section 4 concludes the paper.

# 2. Methodology

## 2.1 Overview of the Proposed PINN-EMFNet

Figure 2 illustrates the overall framework of the proposed method, which consists of two main components: the HA-Encoder and the MSFR-Decoder**.** The method first inputs the raw image into the HA-Encoder network for feature extraction to obtain a rough segmentation result. Then, in the MSFR-Decoder network, a dynamic fusion of multi-scale features is performed using a structure that combines a multi-scale supervision mechanism and a refinement module, refining the preliminary multi-scale segmentation results produced by the encoder. The final segmentation result is obtained from the output at $X^{0,6}$ from the decoder.

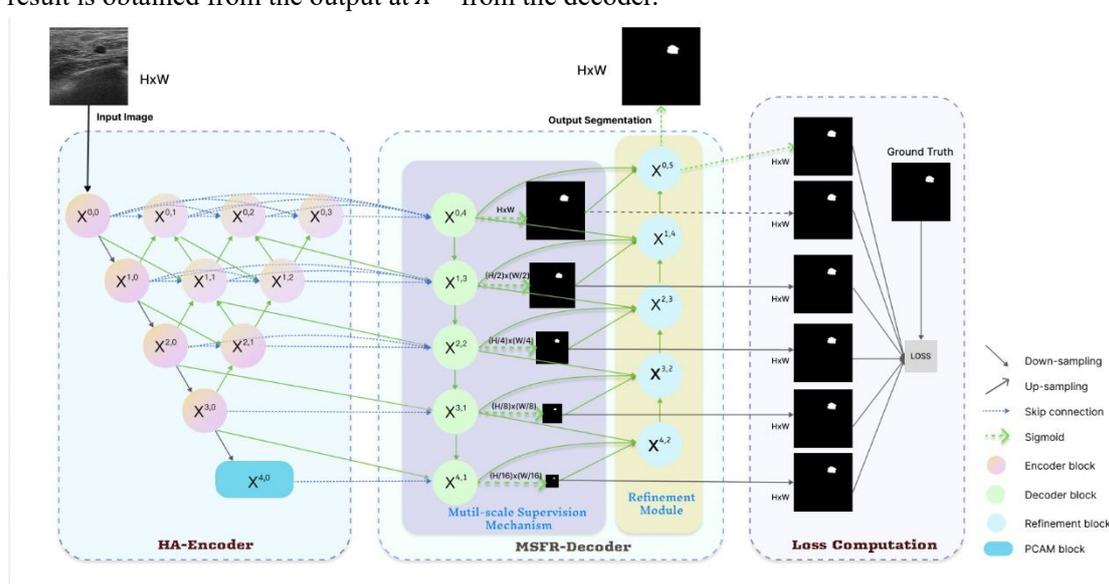

Figure 2. The Architecture of PINN-EMFNet

Specifically, the proposed method is built upon the UNet++ network structure, which has been deeply improved and optimized. In the HA-Encoder section, a cross-structure is employed, enabling network nodes to more comprehensively perceive features at different scales and enhancing the ability to fuse and express multi-scale features. This improved design increases the network's adaptability to complex scenarios and the precision of feature extraction. Additionally, the HA-Encoder section introduces dilated convolutions, effectively expanding the receptive field of the convolutional kernels, thereby capturing complex global and local features more precisely. The application of dilated convolutions allows the network to gain a larger receptive field without increasing the number of parameters, improving the richness and diversity of feature representation. At the last layer of the HA-Encoder, the newly proposed PCAM module is embedded, which optimizes the semantic relationship representation within the image through its superior global modeling capabilities. The HA-Encoder, through multiple structural innovations and the introduction of new modules, significantly improves the network's ability to fuse and express multi-scale features, enhancing the precision of both global and local feature modeling. This enables the encoder to exhibit outstanding performance in suppressing noise interference, capturing tumor boundary details, and improving segmentation reliability.

In the MSFR-Decoder section, the multi-scale supervision mechanism integrates the initial segmentation information generated by the backbone network at coarse scales and dynamically refines the fine-scale features. This mechanism explores adjacent decisions to correct errors that may be caused by image damage, effectively suppressing speckle noise. Furthermore, by integrating segmentation information at different scales, the fine-grained accuracy of the segmentation results is significantly improved. We also introduce a refinement module that merges features at different scales, along with features activated by the sigmoid function, further enhancing the segmentation accuracy. The introduction of the refinement module further optimizes the fusion of cross-scale features, allowing the decoder to fully utilize multi-level feature information. With this design, the decoder excels in enhancing boundary precision and overall segmentation performance, providing strong support for achieving highly reliable and accurate ultrasound image segmentation.

Finally, to further enhance the model's learning capability and segmentation precision, this study innovatively incorporates a physics-informed neural network mechanism in the loss function. PINN combines physical constraints with neural network training, integrating prior knowledge into the deep learning process. This effectively improves the model's ability to consistently model the physical properties of the target region. In the task of ultrasound image segmentation, the PINN mechanism explicitly expresses the physical constraints of the target region, not only improving the accuracy of boundary segmentation but also enhancing the model's robustness against noise and artifacts. As a result, it significantly strengthens the reliability and physical consistency of the segmentation results.

The following sections, 2.2, 2.3, and 2.4, will provide a detailed introduction to the Hierarchical Aggregation Encoder, Multi-Scale Feature Refinement Decoder, and the Loss Function, respectively.

## 2.2 Hierarchical Aggregation Encoder

As mentioned earlier, the HA-Encoder network takes the raw image as input and performs feature extraction to generate a rough segmentation result. The HA-Encoder network proposed in this study is built upon the UNet++ architecture. As shown in Figure 3, the UNet++ network follows a typical encoder-decoder architecture. The innovation of UNet++ lies in its dense skip connections, which establish multiple links between each layer of the encoder and decoder, thereby enhancing the sharing of information across different levels of features. Each skip connection not only directly passes low-level features to higher levels but also enhances cross-level information transfer through nested connections.

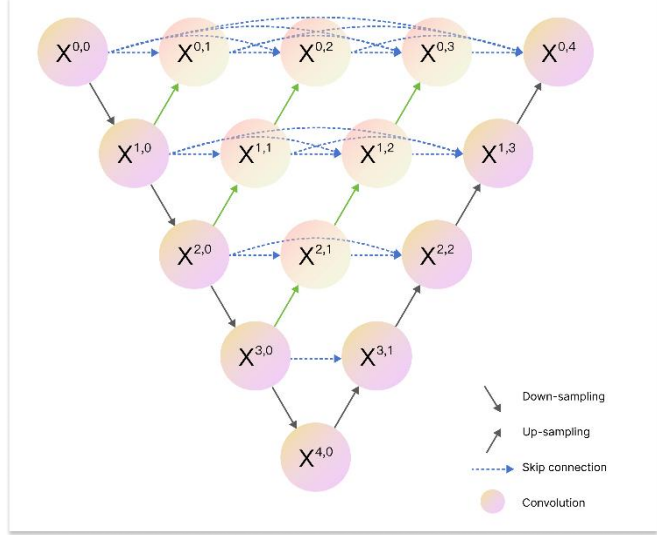

Figure 3. The network of UNet++

Based on the description of UNet++, this study has incorporated a cross-structure at the network nodes to enable more comprehensive perception of features at different scales, thereby enhancing the model's ability to capture multi-scale information. The specific design of the cross-structure is illustrated in Figure 4. In the cross-structure, the node $X^{i+1,j+1}$ receives the feature input from node $X^{i,j}$ while node $X^{i,j+1}$ receives the feature input from node $X^{i+1,j}$

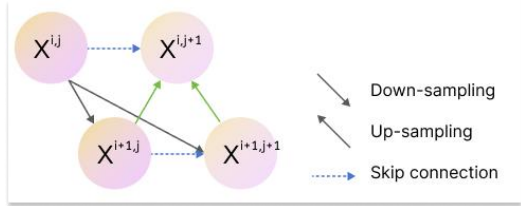

Figure 4. Cross-Structure

To further enhance the feature extraction capability of the model, dilated convolution is introduced into the organizational architecture of the HA-Encoder network. As shown in Figure 5, dilated convolution effectively expands the receptive field by introducing holes (dilation) in the convolution kernel, thereby enabling the network to capture a broader range of contextual information without increasing computational complexity. Dilated convolution assists the model in better capturing long-range dependencies while maintaining computational efficiency. In each layer of the encoder, the convolution operation is replaced with dilated convolution, and the formula is as follows:

$$F_i = \text{Conv}_{\text{dilated}}(F_{i-1}; W_i, d) \quad (1)$$

where $d$ is the dilation rate, typically set to 2 or greater, indicating the extent of the expansion of the convolution kernel.

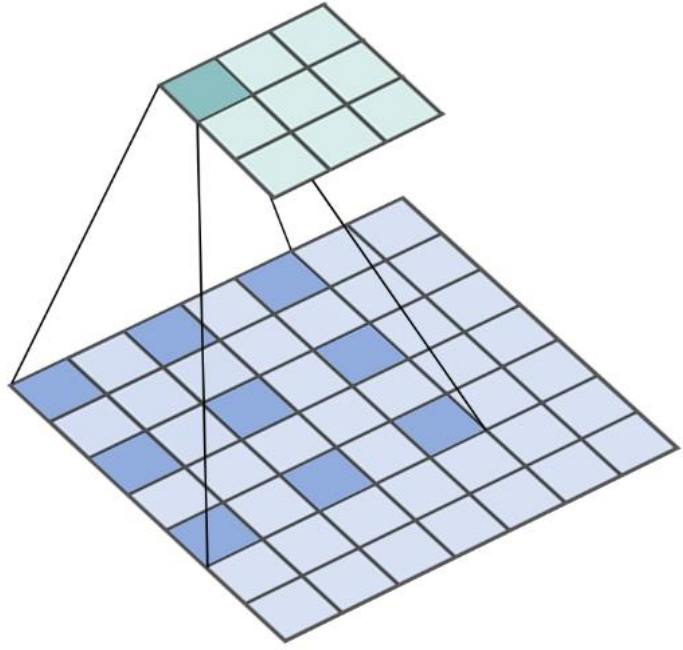

Figure 5. Dilated Convolution

Based on the previously described Encoder structure, this study designs the PCAM module (Parallel Convolutional Attention Module) at the end of the Encoder to enhance the modeling ability of global information and the precision of detail segmentation. The PCAM module is particularly adept at capturing both the global context and fine-grained details of local regions in the image, making it especially effective in dealing with complex backgrounds and detailed region segmentation in medical images.

The core mechanism of the PCAM module integrates adaptive global modeling and local detail capture. It dynamically integrates the global and local relationships between each pixel and other pixels, enabling the model to express the dependencies between features more comprehensively. In terms of implementation, the PCAM module employs an adaptive modeling structure, utilizing effective global context integration and positional encoding methods to process and express the feature map more finely. This allows the model to capture global patterns while maintaining sensitivity to local details. The specific description is as follows:

The PCAM module utilizes multiple parallel convolutional paths, which can effectively extract diverse features from the input image. Each path includes the following operations:

- **Multi-Parallel Convolutional Module**

The multi-parallel convolutional module aims to simultaneously extract diverse feature information through multiple independent convolution paths. Specifically, the input feature map $\mathbf{X} \in \mathbb{R}^{C_{in} \times H \times W}$ is first distributed to $\mathbf{N}$ parallel convolution sub-modules, each of which consists of a convolution layer, a ReLU activation function, and a batch normalization layer. The output feature map of each sub-module is represented as:

$$\mathbf{F}_i = \text{BatchNorm}\left(\text{ReLU}(\text{Conv}(\mathbf{X}))\right), i = 1, 2, \ldots, N \qquad (2)$$

where $\text{Conv}$ represents the convolution operation, and the output channel number is $\text{Conv}/N$.

By processing in parallel, different sub-modules can extract information from different receptive fields and feature scales, thereby enhancing the model's ability to capture complex image features.

- **Global Feature Aggregation**

Based on the output feature maps from the previous multi-parallel convolution module, those maps need to be concatenated along the channel dimension to form a combined feature map. Subsequently, a 1×1 convolutional layer is used for global feature aggregation. To expand the receptive field and enhance the contextual information of the features, a 3×3 dilated convolution is introduced in the module.

$$\mathbf{F}_{ctx} = Conv_{1x1}\left(Conv_{3x3,d=2}(Concat(\mathbf{F}_1, \mathbf{F}_2, \ldots, \mathbf{F}_N) \in \mathbb{R}^{C_{out} \times H \times W})\right) \tag{3}$$

The main function of the global aggregation operation is to perform a linear combination of features from different parallel paths, thereby integrating the feature information and unifying the dimensions. This reduces computational complexity and enhances the comprehensiveness of feature representation. The introduction of dilated convolution ($Conv_{3x3,d=2}$ refers to a 3×3 convolution operation with a dilation rate of 2) effectively expands the receptive field, allowing the model to capture a broader range of contextual information without significantly increasing computational cost.

- **Attention Mechanism**

To further enhance important features and suppress irrelevant ones, an attention mechanism module is designed within the architecture. The specific process is as follows:

$$\mathbf{A} = \sigma\left(Conv_{1x1}\left(\text{ReLU}(Conv_{1x1}(\mathbf{F}_{ctx}))\right)\right) \tag{4}$$

First, a 1×1 convolution is applied to reduce the number of channels to $C_{out}/4$, followed by a ReLU activation function. Then, another 1×1 convolution restores the number of channels back to $C_{out}$. Finally, a Sigmoid function $\sigma$ generates the attention weights $\mathbf{A} \in \mathbb{R}^{C_{out} \times H \times W}$.

These attention weights adaptively adjust the importance of each channel, enabling dynamic re-weighting of the feature map, highlighting key features, and suppressing redundant information.

- **Feature Re-weighting**

Finally, the attention weights are element-wise multiplied with the context-fused features to obtain the final output:

$$\mathbf{Y} = \mathbf{F}_{ctx} \odot \mathbf{A} \tag{5}$$

where, ⊙ denotes the element-wise multiplication operation. This step, by introducing attention weights, allows adaptive adjustment of the different channels in the feature map, thereby enhancing the model's sensitivity to important features and its ability to express them.

Through the synergistic effect of multiple parallel paths, global aggregation, attention mechanisms, and feature reweighting, the visual feature extraction and representation capabilities are significantly enhanced. This visual PCAM module design not only strengthens the model's ability to capture complex image features but also maintains computational efficiency, enabling the extraction of diverse features from the input image and focusing on different types of textures, edges, and structures.

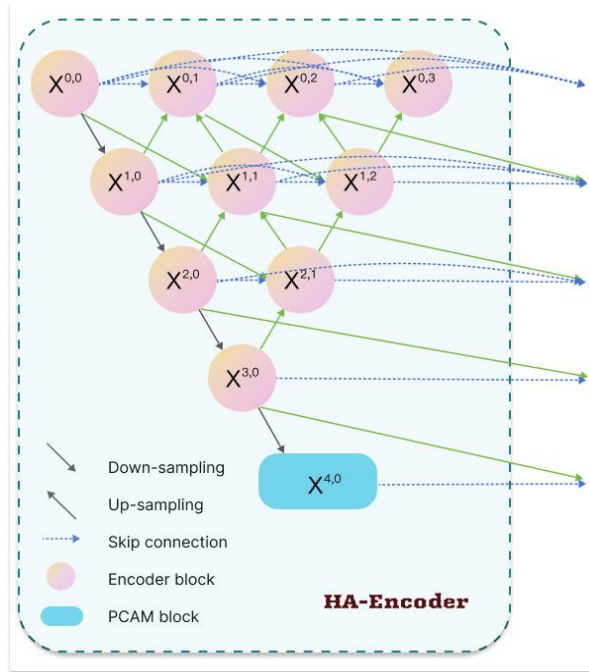

Figure 6 the architecture of Hierarchical Aggregation Encoder

In summary, as illustrated in Figure 6, the HA-Encoder proposed in this study significantly enhances the network's ability to fuse and represent multi-scale features, as well as improving the accuracy of global and local feature modeling through several structural innovations and the introduction of new modules. The design of the cross-structure optimizes information exchange between features at different scales, while dilated convolutions expand the receptive field while maintaining parameter efficiency. The PCAM module further strengthens the modeling of semantic associations, enabling precise segmentation of complex targets in ultrasound images. Overall, this encoder demonstrates excellent performance in suppressing noise interference, capturing tumor boundary details, and improving segmentation reliability, providing an efficient and robust solution for ultrasound image segmentation.

## 2.3 Multi-Scale Feature Refinement Decoder

In the proposed method, the Multi-Scale Feature Refinement Decoder (MSFR-Decoder) adopts a layer-by-layer upsampling strategy. At each layer, operations such as convolution progressively restore the low-resolution feature maps to higher resolutions. The MSFR-Decoder comprises multiple decoding blocks, each consisting of convolution operations, activation functions, and similar components. At each stage, low-resolution features are gradually up-sampled to the target resolution. Different levels of feature maps are fused through cross-layer connections, allowing the network to leverage features at multiple scales and thereby enhancing the model's segmentation capability for complex structures.

As illustrated in Figure 7, the proposed MSFR-Decoder includes two parts: the MSFR-Decoder

itself and the Loss Computation module. Within the MSFR-Decoder, the purple-shaded region denotes a Multi-scale Supervision Mechanism, enabling the segmentation network to receive supervision across multiple feature scales. After each down-sampling, features from corresponding cross-layer connections are fused to generate the output at that scale, with each output corresponding to a particular stage of the decoder's feature resolution. By applying a loss function (detailed in the subsequent subsection) to each multi-scale segmentation output, the model is explicitly guided to learn semantic information and spatial details at various scales. This mechanism not only accelerates network convergence but also further enhances the model's generalization ability in complex scenarios, especially in capturing boundary details. Through this multi-scale output supervision mechanism, the network achieves a better balance between global and local information, thereby significantly improving overall segmentation performance.

Simultaneously, in the proposed method, a Refinement Module is designed in the yellow-shaded area to the right side of the MSFR-Decoder structure. By integrating multi-scale features with features activated by the Sigmoid function, this refinement module enhances segmentation accuracy. Employing down-sampling, up-sampling, skip connections, and Sigmoid activation in combination, this module effectively transmits and refines the features.

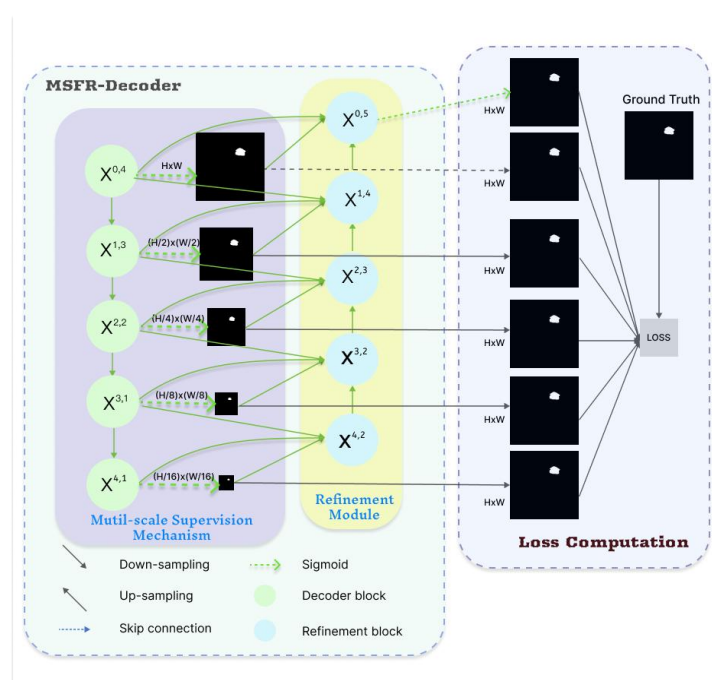

Figure 7 The architecture of Multi-Scale Feature Refinement Decoder

The multi-level feature connections and progressive up-sampling in the MSFR-Decoder enable the model to capture global semantic information while preserving rich boundary details. Through the multi-output deep supervision of the multi-scale supervision mechanism, the model attains a more favorable balance between global and local information, thereby significantly improving overall performance, particularly in complex scenarios and boundary detail delineation. Meanwhile, the refinement module, by integrating multi-scale features and features activated via the Sigmoid function, further enhances segmentation accuracy. This module strengthens the selection of useful information and suppresses irrelevant background, thereby boosting the decoder's performance and robustness.

## 2.4 Loss function

In medical image segmentation tasks, accurately delineating target regions is crucial. However, medical images often face challenges such as class imbalance, complex structures, and image noise, which can lead to suboptimal performance of traditional loss functions in segmentation tasks. Therefore, there is a need to design a loss function that comprehensively addresses pixel-level accuracy, class imbalance, noise suppression, and image smoothness. To address these issues, recent studies have proposed a loss function that integrates Binary Cross-Entropy Loss (BCEWithLogitsLoss)[37, 38], Soft Dice Loss[39], and physical constraints (Total Variation, TV). BCEWithLogitsLoss provides stable and fine-grained pixel-level classification guidance, ensuring that each pixel is accurately classified for precise boundary detection. Soft Dice loss effectively mitigates class imbalance issues by directly optimizing the overlap between predictions and true labels, thereby improving overall segmentation quality. The introduced **TV** constraint promotes the smoothness of segmentation results, reduces noise and discontinuity, and ensures that the segmentation outcomes are more consistent with the physical priors and structural characteristics of medical images. By leveraging the advantages of these three loss functions, the loss design in this study achieves a well-balanced trade-off between pixel-level accuracy, overall segmentation quality, and result smoothness, thereby significantly enhancing the model's robustness and precision in complex medical imaging environments.

BCEWithLogitsLoss is a widely used loss function for binary classification problems, particularly suitable for binary predictions in segmentation tasks. It internally combines the Sigmoid activation function and Binary CrossEntropy loss, thereby eliminating the need to manually apply Sigmoid during training. Its formula is:

$$\mathcal{L}_{\text{BCE}} = -\frac{1}{N} \sum_{i=1}^{N} [y_i \log(\sigma(\hat{y}_i)) + (1 - y_i) \log(1 - \sigma(\hat{y}_i))] \tag{6}$$

where, $\hat{y}_i$ represents the model-predicted segmentation result, $y_i$ denotes the actual label, and $\sigma(\hat{y}_i)$ is the Sigmoid activation function applied to the predicted value.

This loss term drives the model's output to be closer to the true labels, thereby enhancing segmentation accuracy. In medical image segmentation, particularly when dealing with ultrasound images, segmentation precision is often challenged by the irregular morphology of target regions and high noise interference. The advantage of BCEWithLogitsLoss lies in its effective handling of class imbalance issues, especially when the foreground (e.g., lesion areas) constitutes a relatively small proportion and the background dominates. By maximizing the similarity between predictions and true labels, BCEWithLogitsLoss helps the model focus on accurately segmenting the target regions, thereby improving segmentation accuracy.

Soft Dice loss is a loss function commonly used to address class imbalance problems by measuring the overlap between the model's predictions and the true labels to optimize model performance. Compared to traditional cross-entropy loss, Soft Dice loss places greater emphasis on the foreground regions, which is particularly important in medical imaging. This is because, in most medical images (including ultrasound images), the target regions (such as lesions, tumors, or organs) occupy a much smaller proportion than the background areas. The computation of the Soft Dice coefficient is as follows:

$$\mathcal{L}_{\text{Dice}} = 1 - \frac{2\sum_{i=1}^{N}\hat{y}_i y_i}{\sum_{i=1}^{N}\hat{y}_i + \sum_{i=1}^{N} y_i} \tag{7}$$

where, $\hat{y}_i$ and $y_i$ represent the predicted and true pixel values, respectively.

Dice loss optimizes the model by maximizing the overlap between predictions and true labels, making it particularly suitable for tasks with imbalanced classes. In ultrasound image segmentation, because target regions are usually small and irregularly shaped, traditional loss functions may not effectively handle such imbalance. Soft Dice loss alleviates this problem by maximizing the overlap between target and predicted regions.

To further enhance the smoothness of the model's segmentation results and reduce boundary noise, this study introduces Total Variation loss as a physical constraint. Total Variation is a regularization method commonly used in the field of image processing. It suppresses noise by penalizing gradient variations in the image while preserving boundary information. This work leverages TV loss to constrain the predicted segmentation images, ensuring that the segmentation outcomes are smoother and free from overly rough or irregular boundaries. Specifically, TV loss measures the smoothness of the image by computing the gradients of the predicted segmentation image in the $x$ and $y$ directions. The calculation of TV loss is based on the gradients of the predicted image, as follows:

$$\mathcal{L}_{\text{TV}} = \frac{1}{N}\sum_{i=1}^{N}\left(\left|\frac{\partial \hat{y}_i}{\partial x}\right| + \left|\frac{\partial \hat{y}_i}{\partial y}\right|\right) \tag{8}$$

where, $\hat{y}_i$ represents the model's predicted output, and $\frac{\partial \hat{y}_i}{\partial x}$ and $\frac{\partial \hat{y}_i}{\partial y}$ denote the gradients of the predicted image in the $x$ and $y$ directions, respectively.

By penalizing the L1 norm of these gradients, this loss term helps reduce noise and irregularities in the segmentation boundaries while preserving the continuity of edges, thereby preventing excessive smoothing. By incorporating the physical constraints of Physics-Informed Neural Networks into the loss function, this study achieves a better balance between segmentation accuracy and result smoothness. This study opts to use the gradients of the predicted image rather than those of the input image because the focus is on the smoothness of the segmentation results rather than the inherent smoothness of the input images. By computing the gradients of the predicted segmentation maps, regions with significant gradient changes—typically corresponding to segmentation boundaries—can be identified and optimized. Minimizing the TV loss helps maintain the smoothness of these boundaries, thereby enhancing the quality of the segmentation outcomes.

To integrate the aforementioned losses, this study defines the total loss function $\mathcal{L}_{\text{total}}$ as a weighted sum of BCEWithLogitsLoss, Soft Dice Loss, and Total Variation loss. The specific loss function can be expressed as:

$$\mathcal{L}_{\text{total}} = \lambda_0 \mathcal{L}_{\text{BCE}} + \lambda_1 \mathcal{L}_{\text{TV}} + \lambda_2 \mathcal{L}_{\text{BCE}}$$

$\lambda_0$, $\lambda_1$ and $\lambda_2$ are hyperparameters that balance the various loss components, used to control the contribution of each loss term. Through this weighted loss function, the study can simultaneously optimize the model's segmentation accuracy and the smoothness of the segmentation results.

During the training process, the optimizer minimizes the total loss function $\mathcal{L}_{\text{total}}$ to achieve optimal segmentation performance. The loss function proposed in this study integrates Binary Cross-Entropy loss, Soft Dice loss and Total Variation loss from PINN, collectively driving the model to produce high-quality segmentation results. By incorporating BCEWithLogitsLoss, this study addresses class imbalance and prediction accuracy issues; Soft Dice loss enhances the model's sensitivity to target regions, particularly those that are small and irregularly shaped. Finally, leveraging the physical constraints of PINN, this study is able to smooth the segmentation results and reduce the impact of noise, thereby obtaining more precise and smooth segmentation outcomes and improving the model's performance in complex medical image segmentation tasks.

# 3. Experiments and Results

## 3.1 Data Information

In this study, the proposed novel segmentation network was comprehensively evaluated using two publicly available breast ultrasound lesion datasets, BUSIS and BUSI. These datasets were collected from different regions and time periods and were annotated by professional radiologists to delineate lesion boundaries, ensuring high quality and reliability. They are widely utilized in research on breast ultrasound image segmentation tasks. Below is a detailed description of the two datasets:

BUSIS originates from the UDIAT Diagnostic Center of Tau'lI Company in Sabadell, Spain, and was collected in 2012 using the Siemens ACUSON Sequoia C512 system with a 17L5 high-definition linear array transducer (8.5 MHz) to acquire breast ultrasound images. This dataset comprises 163 breast ultrasound images, each obtained from different female subjects and containing a single annotated lesion region. The average resolution of the images is 760×570 pixels. The dataset includes 110 benign cases, encompassing 39 fibroadenomas, 65 unspecified cysts, and 6 other types of lesions. Malignant cases total 53, divided into 2 invasive lobular carcinomas, 4 ductal carcinoma in situ, 40 invasive ductal carcinomas, and 7 other unspecified malignant lesions. The boundaries of the breast lesions were manually annotated by experienced radiologists, providing high-quality segmentation masks for image segmentation tasks.

BUSI is sourced from the Women's Early Detection and Treatment Center for Cancer at Baheya Hospital in Cairo, Egypt, and was collected in 2018. It includes data from 600 patients aged between 25 and 75 years. The images were acquired using LOGIQ E9 ultrasound devices and the LOGIQ E9 Agile ultrasound system with an ML6-15-D matrix linear probe (10-5 MHz), resulting in 780 breast ultrasound images with an average resolution of 500×500 pixels. The images in this dataset are categorized into three classes: 133 normal images, 437 benign lesions, and 210 malignant lesions. Each image's lesion region was manually annotated by radiologists, and the lesion boundaries were generated into segmentation masks using MATLAB tools. BUSI provides a substantial sample size and high-quality annotations, offering robust support for breast ultrasound classification and segmentation tasks.

The combination of these two datasets provides a rich source of training and testing data for the development and evaluation of the breast ultrasound segmentation network. BUSIS offers detailed information on lesion type distribution and higher resolution images, while BUSI covers a larger sample size and a broader age range of patients (25-75 years). By utilizing these high-quality and diverse datasets, the proposed network can validate its performance across varied scenarios, ensuring its robustness and reliability in practical applications.

## 3.2 Experimental Settings

### 3.2.1 Baseline Method

The novel breast cancer ultrasound image segmentation network proposed in this study was developed using Python 3.12, CUDA 12.1, and PyTorch 2.2.2 on a Windows 11 workstation equipped with an NVIDIA GeForce GTX 4090D GPU (24.0 GB). The network was trained for 450 epochs using the Adam optimizer.

To comprehensively evaluate the performance of the proposed method, five advanced medical image segmentation models were compared: convolutional networks (U-Net, UNet++, Att U-Net, MDF-Net) and transformer-based networks (HiFormer-L and TransUnet). In the experiments, a five-fold cross-validation strategy was employed for each of the two datasets. Specifically, in each fold, four-fifths of the data were used for training, and the remaining one-fifth served as the test set. All benign and malignant images were evenly distributed across each fold to ensure balanced data distribution. This cross-validation approach effectively reduces randomness in model performance evaluation and enhances the reliability and robustness of the results.

In terms of data preprocessing, all images were normalized before being input into the network to minimize the impact of intensity variations between samples. This normalization process not only aids in accelerating the convergence speed of network training but also enhances the model's adaptability to different samples.

### 3.2.2 Evaluation metrics:

For performance evaluation, the segmentation results of the models were reported as the mean ± standard deviation on the test sets, thereby providing insights into the central tendency and variability of segmentation performance. To comprehensively measure the segmentation performance of different methods, this study utilized the following key evaluation metrics:

- Dice Similarity Coefficient (DSC): Measures the degree of overlap between the predicted results and the true segmentation, used to evaluate the overall accuracy of the segmentation results. DSC ranges from [0, 1], where 1 indicates a perfect segmentation and 0 indicates no overlap.

- Intersection over Union (IoU): Calculates the ratio of the intersection to the union of the predicted results and the true segmentation, serving as another important metric to measure segmentation accuracy. IoU also ranges from [0, 1], with values closer to 1 indicating greater consistency between the predicted and true segmentation regions.
- Sensitivity (SE): Also known as recall, reflects the model's ability to detect positive classes (e.g., lesion areas). It is defined as the proportion of true positives out of all actual positives. Sensitivity ranges from [0, 1], with higher values indicating better detection of positive classes.
- Specificity (SP): Measures the model's ability to identify negative classes (e.g., non-lesion areas). It is defined as the proportion of true negatives out of all actual negatives. Specificity ranges from [0, 1], with higher values indicating better detection of negative classes.

The formulas for Dice Similarity Coefficient and Intersection over Union are as follows:

$$DSC = \frac{2|P \cap G|}{|P| + |G|}$$

$$IoU = \frac{|P \cap G|}{|P \cup G|}$$

$$SE = \frac{TP}{TP + FN}$$

$$SP = \frac{TN}{TN + FP}$$

where:
- $P$ denotes the Predicted region,
- $G$ denotes the Ground truth region,
- True Positive (TP): The number of samples correctly predicted as positive.
- True Negative (TN): The number of samples correctly predicted as negative.
- False Positive (FP): The number of samples incorrectly predicted as positive.
- False Negative (FN): The number of samples incorrectly predicted as negative.

Through these metrics, this study can comprehensively analyze and compare the models' performance in lesion segmentation tasks from various perspectives, thereby validating the effectiveness and robustness of the proposed improvement methods.

## 3.3 Quantitative Comparison with State of the Art Methods

In this section of the experiment, this study comprehensively evaluates the performance of the PINN-EMFNet method in the tumor segmentation task. Specifically, the study compares it with five state-of-the-art medical image segmentation models on the BUSIS and BUSI datasets, including classical convolutional neural networks (such as U-Net, U-Net++, Att U-Net, MDF-Net) and Transformer-based networks (such as HiFormer-L and TransUnet). Through these comparative experiments, the study aims to thoroughly analyze the advantages of PINN-EMFNet in the tumor segmentation task, particularly in the segmentation of different types of breast tumors (benign and malignant). The following sections will provide a detailed discussion of the comparative results of each model across different evaluation metrics.

From the results presented in Table 1, various segmentation methods exhibit differing performances across different metrics on the BUSIS dataset. Notably, the PINN-EMFNet proposed in this study significantly outperforms other algorithms in segmentation tasks for different types of breast masses (benign and malignant).

In terms of the Dice Similarity Coefficient, PINN-EMFNet demonstrates the best performance across all types of breast mass segmentation, particularly for benign masses ($87.33 \pm 4.89$) and malignant masses ($79.93 \pm 6.05$), with an overall DSC of $87.62 \pm 4.81$. This performance is markedly superior to classic models such as U-Net ($83.24 \pm 4.54$) and HiFormer-L ($81.45 \pm 5.97$). Additionally, compared to MDF-Net ($85.07 \pm 3.81$), PINN-EMFNet shows a noticeable improvement in DSC, showcasing its robust adaptability in handling the complex morphology of breast masses.

Regarding the Intersection over Union metric, PINN-EMFNet achieves segmentation accuracies of $79.73 \pm 5.28$ for benign masses and $70.55 \pm 5.09$ for malignant masses, with an overall IoU of $80.10 \pm 8.11$. Although this performance is comparable to MDF-Net's benign mass segmentation ($79.08 \pm 5.44$), PINN-EMFNet exhibits more stable performance in the segmentation of malignant masses. When compared to methods such as U-Net, U-Net++, and TransUnet, PINN-EMFNet shows a significant advantage in the segmentation of malignant masses, indicating its superior discriminative capability in handling the differentiated features of various lesion types.

The Sensitivity and Specificity of PINN-EMFNet also perform exceptionally well, with SE values of $84.82 \pm 3.33$ and SP of $99.49 \pm 0.22$. Specifically, in terms of sensitivity, PINN-EMFNet significantly outperforms U-Net ($81.48 \pm 2.52$) and Att U-Net ($81.11 \pm 4.32$), indicating a lower false positive rate in detecting breast masses and demonstrating higher detection capabilities.

The primary advantages of PINN-EMFNet lie in its higher accuracy and stability in distinguishing between benign and malignant masses, coupled with a lower standard deviation, which indicates consistent segmentation results across different samples. Compared to traditional or transformer-based networks such as U-Net, U-Net++, Att U-Net, HiFormer-L, and TransUnet, PINN-EMFNet integrates effective feature fusion and enhancement modules, providing greater flexibility and precision in fine-grained segmentation tasks. Furthermore, PINN-EMFNet maintains high specificity while also enhancing sensitivity, which is crucial for the early screening of breast cancer. High specificity ensures that the model reduces false positives, thereby decreasing the psychological burden on patients, while high sensitivity indicates that the model is highly adept at lesion detection, effectively minimizing missed diagnoses.

TABLE 1 COMPARISON OF EVALUATION METRICS ON BUSIS DATASET

| Method | Mass Type | BUSIS | | | |
| --- | --- | --- | --- | --- | --- |
| | | DSC | IoU | SE | SP |
| U-Net | Benign | 84.15±2.21 | 76.00±2.68 | 86.07±3.97 | 99.52±0.20 |
| | Malignant | 76.11±6.05 | 66.65±7.46 | 77.48±10.39 | 99.12±0.50 |
| | All | 81.48±2.52 | 72.90±3.11 | 83.24±4.54 | 99.39±0.22 |
| U-Net++ | Benign | 84.89±1.66 | 77.19±2.25 | 86.46±3.60 | 99.58±0.15 |
| | Malignant | 75.15±7.22 | 65.58±7.74 | 75.27±10.31 | 99.06±0.53 |
| | All | 81.66±2.72 | 73.33±3.33 | 82.76±3.84 | 99.40±0.20 |
| Att U-Net | Benign | 84.20±2.98 | 75.97±3.85 | 85.71±4.46 | 99.50±0.43 |
| | Malignant | 74.89±8.75 | 65.11±9.73 | 76.15±12.96 | 98.97±0.55 |
| | All | 81.11±4.32 | 72.37±5.18 | 82.55±6.57 | 99.32±0.40 |
| HiFormer-L | Benign | 83.81±3.90 | 75.32±5.04 | 83.55±5.86 | 99.64±0.28 |

|  | Malignant | 76.91±9.58 | 66.44±9.24 | 77.19±14.10 | **99.17±0.54** |
|  | All | 81.52±3.55 | 72.38±3.77 | 81.45±5.97 | 99.49±0.27 |
| TransUnet | Benign | 84.66±3.64 | 76.45±4.07 | 86.25±5.06 | 99.47±0.40 |
|  | Malignant | 76.21±8.48 | 65.29±8.64 | 76.64±9.83 | 99.06±0.53 |
|  | All | 81.85±3.98 | 72.74±4.22 | 83.07±3.94 | 99.33±0.42 |
| MDF-Net | Benign | 85.90±1.15 | 78.49±1.88 | 87.27±3.14 | 99.59±0.28 |
|  | Malignant | 79.08±5.44 | 70.49±7.01 | **80.60±7.79** | 99.09±0.44 |
|  | All | 83.63±2.30 | 75.83±3.02 | 85.07±3.81 | 99.42±0.27 |
| PINN-EMFNet | Benign | **87.33±4.89** | **79.93±6.05** | **87.62±4.81** | **99.69±0.28** |
|  | Malignant | **79.73±5.28** | **70.55±5.09** | 80.10±8.11 | 99.09±0.79 |
|  | All | **84.82±3.33** | **76.83±4.14** | **85.15±3.40** | 99.49±0.22 |

Based on the segmentation metric results in TABLE 2 for the BUSI breast ultrasound image dataset, PINN-EMFNet demonstrates significant advantages in the segmentation performance of breast masses compared to other comparative methods.

PINN-EMFNet achieved Dice Similarity Coefficient scores of $80.41 \pm 6.00$ for benign masses and $73.51 \pm 6.04$ for malignant masses, with an overall DSC of $83.66 \pm 5.59$. These results are notably superior to those of U-Net (overall DSC of $76.79 \pm 5.44$) and U-Net++ (overall DSC of $75.82 \pm 5.69$), indicating that PINN-EMFNet is better able to match the target regions in the task of breast mass segmentation. Compared to MDF-Net, PINN-EMFNet exhibits a slight disadvantage in the segmentation of malignant masses ($73.51 \pm 6.04$ versus $74.20 \pm 6.56$ for MDF-Net). However, PINN-EMFNet demonstrates more stable performance in the segmentation of benign masses and overall segmentation, particularly improving the overall DSC by nearly one percentage point.

Regarding the Intersection over Union metric, PINN-EMFNet achieved IoU scores of $74.01 \pm 5.84$ for benign masses, $63.82 \pm 5.88$ for malignant masses, and $74.29 \pm 3.67$ overall. Compared to other algorithms, PINN-EMFNet exhibits significant advantages in IoU for benign masses and overall segmentation. Specifically, compared to U-Net (overall IoU of $67.51 \pm 4.23$) and HiFormer-L (overall IoU of $71.94 \pm 6.03$), PINN-EMFNet shows an improvement of approximately 6-7 percentage points. While MDF-Net has similar IoU performance for benign and malignant masses compared to PINN-EMFNet, PINN-EMFNet demonstrates a lower standard deviation, indicating better segmentation stability and more consistent handling of different types of data.

In terms of Sensitivity (SE), PINN-EMFNet achieved an SE value of $78.28 \pm 5.13$ in overall segmentation, which is higher than U-Net ($75.51 \pm 6.04$) and TransUnet ($76.05 \pm 6.41$), indicating that the model has a stronger ability to detect target regions with a lower false negative rate. Regarding Specificity (SP), PINN-EMFNet achieved an SP of $98.17 \pm 1.18$, which is very close to the specificity of other models, such as U-Net++ ($98.47 \pm 1.12$) and Att U-Net ($98.48 \pm 1.04$). This suggests that PINN-EMFNet performs comparably to other models in maintaining good discrimination of background regions, with a slight advantage.

TABLE 2 COMPARISON OF EVALUATION METRICS ON BUSI DATASET

| Method | Mass Type | BUSI | | | |
|---|---|---|---|---|---|
|  |  | DSC | IoU | SE | SP |
| U-Net | Benign | 78.59±6.61 | 71.03±6.79 | 81.40±6.40 | 99.00±0.66 |
|  | Malignant | 69.32±5.85 | 58.35±5.37 | 67.51±4.23 | 97.41±41.65 |
|  | All | 75.51±6.04 | 66.82±6.08 | 76.79±5.44 | 98.47±0.97 |
| U-Net++ | Benign | 79.11±6.79 | 71.59±7.02 | 80.64±6.75 | **99.05±0.69** |
|  | Malignant | 68.37±5.82 | 57.05±5.60 | 66.11±4.59 | 97.32±2.04 |
|  | All | 75.54±6.17 | 66.77±6.29 | 75.82±5.69 | 98.47±1.12 |
| Att U-Net | Benign | 78.21±7.03 | 70.64±7.25 | 81.56±6.96 | 98.92±0.72 |
|  | Malignant | 68.49±6.43 | 57.47±6.02 | 65.89±6.51 | **97.59±1.72** |
|  | All | 74.98±6.43 | 66.27±6.50 | 76.35±6.10 | **98.48±1.04** |
| HiFormer-L | Benign | 77.71±8.69 | 70.22±8.44 | 80.26±9.14 | 98.82±0.87 |
|  | Malignant | 70.67±6.60 | 59.80±6.02 | 71.94±6.03 | 95.47±3.99 |
|  | All | 75.36±7.02 | 66.76±6.85 | 77.49±8.23 | 97.71±1.83 |
| TransUnet | Benign | 78.48±7.81 | 70.56±8.12 | 82.52±8.51 | 98.27±0.48 |
|  | Malignant | 71.18±5.85 | 59.78±5.37 | 72.75±4.92 | 95,32±3.11 |
|  | All | 76.05±6.41 | 66.97±6.53 | 79.27±7.02 | 97.29±1.22 |
| MDF-Net | Benign | **80.93±6.63** | **74.20±6.56** | 82.81±6.73 | 98.99±0.83 |
|  | Malignant | 72.69±4.30 | 62.21±3.74 | 72.67±5.42 | 96.67±2.64 |
|  | All | 78.20±5.72 | 70.22±5.53 | 79.44±5.73 | 98.22±1.36 |
| PINN-EMFNet | Benign | 80.41±6.00 | 73.51±6.04 | **83.66±5.59** | 98.86±0.83 |
|  | Malignant | **74.01±5.84** | **63.82±5.88** | **74.29±3.67** | 96.78±2.00 |
|  | All | **78.28±5.13** | **70.29±5.1** | **80.54±4.45** | 98.17±1.18 |

## 3.4 Comparison of Tumor Segmentation with State of the Art Methods

In this section of the experiment, the study provides a visual comparison of the performance of different algorithms in the ultrasound image tumor segmentation task, with a particular focus on analyzing the segmentation results of PINN-EMFNet and five other state-of-the-art medical image segmentation models on the BUSIS and BUSI datasets. By presenting and analyzing the segmentation results of each algorithm on target regions with varying types, shapes, and complexities, this study aims to thoroughly explore the advantages of PINN-EMFNet in preserving object boundaries, capturing complex morphological features, and detecting small targets. The following sections will offer a detailed comparison of the models in terms of segmentation accuracy and visual quality.

Figure 8 presents a comparison of the predicted results of various algorithms in the tumor segmentation tasks on a subset of ultrasound images from the BUSIS dataset. The first column displays the original ultrasound images, which contain target regions of different types and shapes. The second column shows the Ground Truth. The remaining columns illustrate the segmentation results of each algorithm, including the proposed method, PINN-EMFNet, and several mainstream algorithms: U-Net++, HiFormer-L, MDF-Met, and TransUnet.

From the segmentation results, PINN-EMFNet (second column) exhibits a high degree of similarity to the GT, especially in simpler target regions, such as those in the first, second, and fifth rows of the images. In contrast, other algorithms, such as U-Net++ and HiFormer-L, tend to show deviations at the boundaries of target regions, sometimes failing to accurately capture the complete edges of targets or small targets.

PINN-EMFNet performs exceptionally well in segmenting the edges of targets, particularly in the third and fifth rows of the ultrasound images, successfully capturing more complex boundary contours. In comparison, methods like U-Net++ and HiFormer-L tend to produce smoothed edges under similar conditions, leading to the loss of detail. For targets with larger areas and irregular shapes (as seen in the fourth row of the ultrasound images), PINN-EMFNet demonstrates good target integrity, closely matching the contours of the GT. Meanwhile, the segmentation results of MDF-Met and HiFormer-L show some parts missing. PINN-EMFNet accurately detects and segments small targets, whereas U-Net++ and HiFormer-L have relatively weaker capabilities in capturing small targets, resulting in larger deviations. MDF-Met and TransUnet also exhibit instability when handling such small targets.

Overall, PINN-EMFNet demonstrates strong edge-preserving capabilities and adeptness in capturing targets with complex morphologies in ultrasound image segmentation. Compared to other benchmark methods, PINN-EMFNet achieves results closer to the GT, showing significant advantages in detecting small targets, maintaining complex edges, and handling background noise.

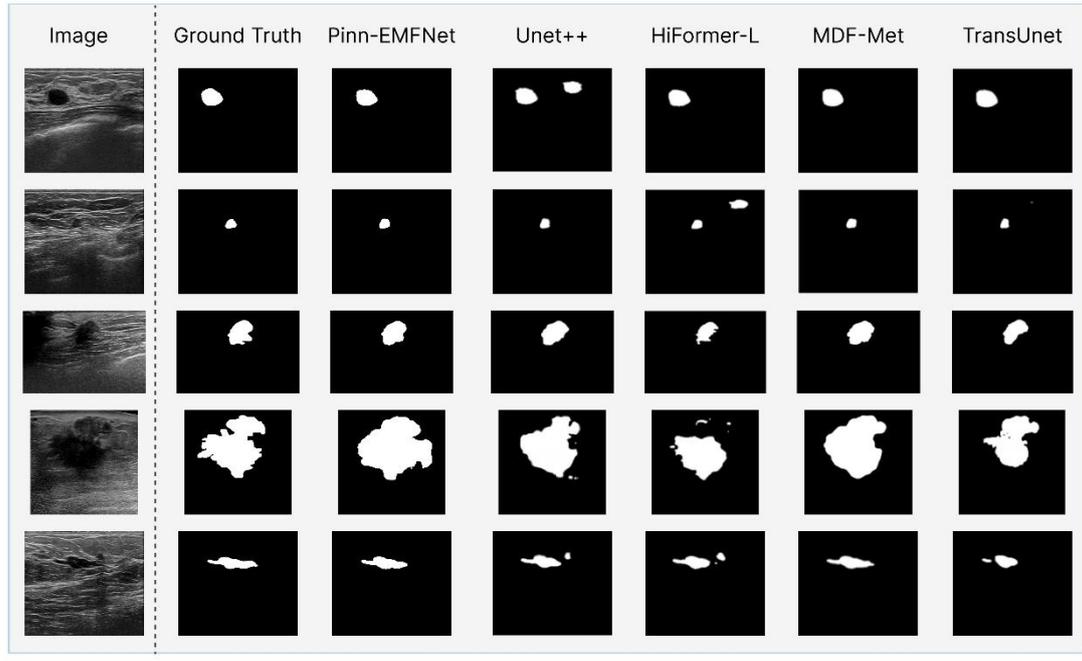

Figure 8 Segmentation results of different methods on BUSIS

Figure 9 illustrates a comparison of the predicted results of various algorithms in the tumor segmentation tasks on a subset of ultrasound images from the BUSI dataset. The other fundamental information remains consistent with the description of Figure 8. Overall, PINN-EMFNet produces segmentation results that are closest to the Ground Truth across all scenarios, particularly excelling in the integrity of target region boundaries and the restoration of target morphology. In contrast, other algorithms, including U-Net++, HiFormer-L, MDF-Met, and TransUnet, exhibit inconsistent performance in segmentation results across certain scenarios, demonstrating shortcomings in handling details, complex edges, and background noise.

PINN-EMFNet shows a distinct advantage in processing details and edges. In the first and fourth rows of the images, which display small target regions, PINN-EMFNet accurately segments the fine details of the small targets. In comparison, U-Net++ and HiFormer-L produce segmentation results with incomplete target boundaries, showing significant instances of missed segmentation in the target areas. Similarly, MDF-Met and TransUnet fail to capture all the details of small targets, resulting in noticeably smaller segmented shapes.

In the second row of the images, where the target region is larger and irregularly shaped with some background noise interference, PINN-EMFNet successfully segments the complete target, closely resembling the GT. Conversely, U-Net++ produces segmentation results with rough edges and missegmented areas. HiFormer-L and MDF-Met are unable to sufficiently suppress background noise during segmentation, leading to blurred edges and even the appearance of artifacts. Although TransUnet captures the general shape of the target, its boundaries are unsmooth, and some details are missing.

For larger and more complexly shaped target regions, such as those in the fifth row of the images, PINN-EMFNet demonstrates optimal target integrity and edge preservation. In contrast, U-Net++ and HiFormer-L show partial loss of target regions and imprecise edges in their segmentation results. MDF-Met and TransUnet also fail to replicate all the edge details of the GT when handling these complex targets, resulting in significant morphological deviations, especially in the detailed representation of complex edges.

In summary, PINN-EMFNet exhibits outstanding performance in handling various types of ultrasound image segmentation tasks, particularly excelling in capturing target details, maintaining edge integrity, and suppressing background noise compared to other comparative methods. The figures demonstrate that PINN-EMFNet can accurately reproduce the target shapes and boundaries of the GT, showcasing superior adaptability and robustness.

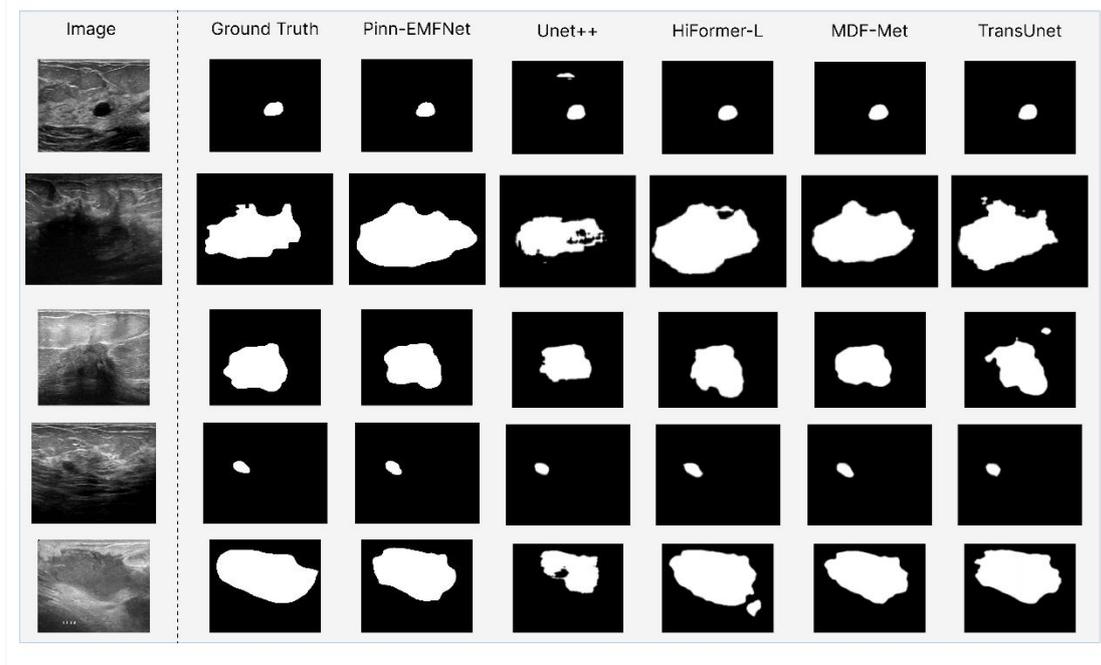

Figure 9 Segmentation results of different methods on BUSI

## 3.5 Quantitative Comparison under Different Scenarios

Based on the results presented in TABLE 3, PINN-EMFNet exhibits varying segmentation performance across the two breast ultrasound image datasets, BUSIS and BUSI. PINN-EMFNet demonstrates superior overall performance on the BUSIS dataset compared to the BUSI dataset, particularly in the two core segmentation metrics: Dice Similarity Coefficient and Intersection over Union. The BUSIS dataset may contain more standardized data or offer a more favorable annotation and feature distribution for model learning, enabling PINN-EMFNet to execute tumor segmentation more effectively.

On the BUSIS dataset, PINN-EMFNet exhibits higher sensitivity and specificity, which are crucial for breast cancer screening—high sensitivity indicates a strong capability to identify tumors, while high specificity reduces false positives, thereby decreasing the psychological burden on patients. Overall, the segmentation results of PINN-EMFNet on the BUSIS dataset are more stable and achieve higher segmentation accuracy for both benign and malignant tumors. This may suggest that the BUSIS dataset provides greater support for model training and validation in terms of data balance and annotation quality.

Conversely, on the BUSI dataset, the segmentation performance of the model experiences a slight decline. This could be attributed to more complex data characteristics or greater challenges in annotation within the BUSI dataset, which may hinder the model's ability to maintain the same level of performance as observed with the BUSIS dataset.

TABLE 3 COMPARISON OF EVALUATION METRICS ON BUSIS AND BUSI DATASETS

| Dataset | Mass Type | PINN-EMFNet | | | |
|---|---|---|---|---|---|
| | | DSC | IoU | SE | SP |
| BUSIS | Benign | 87.33±4.89 | 79.93±6.05 | 87.62±4.81 | 99.69±0.28 |
| | Malignant | 79.73±5.28 | 70.55±5.09 | 80.10±8.11 | 99.09±0.79 |
| | All | 84.82±3.33 | 76.83±4.14 | 85.15±3.40 | 99.49±0.22 |
| BUSI | Benign | 80.41±6.00 | 73.51±6.04 | 83.66±5.59 | 98.86±0.83 |
| | Malignant | 74.01±5.84 | 63.82±5.88 | 74.29±3.67 | 96.78±2.00 |
| | All | 78.28±5.13 | 70.29±5.10 | 80.54±4.45 | 98.17±1.18 |

## 3.6 Ablation Study

## 3.6.1 Comparison of PCAM and Transformer

The Transformer architecture also possesses strong capabilities in capturing long-range dependencies and global contextual information. In this study, this study also experimented with replacing the last layer of the HMSA-Encoder with a Transformer structure. The specific experimental results are shown in TABLE 5.

From TABLE 4, it can be seen that the Transformer structure slightly outperforms the PCAM

structure in the overall Dice Similarity Coefficient, achieving a DSC of 84.71, while the PCAM structure achieved 84.82, with minimal difference between the two. In terms of Intersection over Union, the PCAM structure (76.83) is slightly higher than the Transformer structure (76.49), indicating an advantage in the coverage of segmented regions. For Sensitivity (SE), the Transformer structure achieved 85.72, slightly higher than the PCAM structure's 85.15, with no significant difference between the two. Regarding Specificity (SP), the PCAM structure's performance (99.49) is slightly higher than the Transformer structure's (99.48), indicating that both structures have nearly identical capabilities in eliminating background interference and reducing false positives.

Overall, there is no significant difference in segmentation performance between the two models. The Transformer structure slightly outperforms the PCAM structure in DSC and SE, while the PCAM structure has a slight advantage in IoU and SP. Notably, the PCAM structure achieves better IoU performance in the segmentation of malignant masses, reaching 70.55 compared to the Transformer structure's 69.27, demonstrating its superior coverage in segmenting more complex regions.

In terms of model file size, the Transformer structure has a model file size of 874 MB, whereas the PCAM structure's model file size is 540 MB. The PCAM structure has significantly fewer model parameters than the Transformer structure, indicating more lightweight storage and deployment. Additionally, the Transformer structure is slower than the PCAM structure in both training and inference phases, and it also consumes more operational memory. This situation is typically because the Transformer structure requires greater computational resources to handle the self-attention mechanism. Although it has a stronger feature-capturing capability, it is less resource-efficient compared to the PCAM structure.

The advantages of the PCAM structure in model size, operational efficiency, and memory usage make it more economical and convenient for practical applications, especially in scenarios that require extensive model deployment and rapid inference. While the Transformer slightly outperforms the PCAM structure in certain metrics, its higher resource consumption and slower inference speed limit its applicability in large-scale deployments. Therefore, in application scenarios that require a balance between performance and resources, the PCAM structure may be a more suitable choice, whereas the Transformer structure is more appropriate for scenarios with extremely high performance requirements and relatively abundant computational resources. Maintaining the accuracy of mathematical terminology and markup format is crucial.

TABLE 4 COMPARISON OF EVALUATION METRICS USING PCAM AND TRANSFORMER IN PINN-EMFNET ON BUSIS DATASET

| Method | Mass Type | BUSIS | | | |
|---|---|---|---|---|---|
| | | DSC | IoU | SE | SP |
| PINN-EMFNet (PCAM) | Benign | 87.33±4.89 | 79.93±6.05 | 87.62±4.81 | 99.69±0.28 |
| | Malignant | 79.73±5.28 | 70.55±5.09 | 80.10±8.11 | 99.09±0.79 |
| | All | 84.82±3.33 | 76.83±4.14 | 85.15±3.40 | 99.49±0.22 |
| PINN-EMFNet (Transformer) | Benign | 87.37±3.99 | 80.09±4.89 | 88.48±4.26 | 99.65±0.25 |
| | Malignant | 79.37±6.64 | 69.27±7.29 | 80.12±7.79 | 99.12±0.40 |
| | All | 84.71±2.84 | 76.49±3.63 | 85.72±2.93 | 99.48±0.15 |

## 3.6.2 Effectiveness of Dilated Convolution

To demonstrate the positive impact of dilated convolutions on the network framework, this study conducted a comparative analysis of two proposed models. One model utilizes dilated convolutions (PINN-EMFNet with dilation), while the other does not (PINN-EMFNet without dilation). TABLE 5 provides a detailed presentation of four primary performance metrics: Dice Similarity Coefficient, Intersection over Union, Sensitivity, and Specificity. The experimental results indicate that the model employing dilated convolutions significantly outperforms the model without dilated convolutions in DSC (overall 84.82 vs. 82.95) and IoU (overall 76.83 vs. 75.19), particularly in the segmentation of malignant masses. Dilated convolutions effectively expand the receptive field, allowing the model to better capture global contextual information without significantly increasing computational complexity, thereby enhancing segmentation accuracy, especially in handling the boundaries of complex lesions. Additionally, the model with dilated convolutions exhibits an advantage in Sensitivity (overall 85.15 vs. 84.59), indicating greater sensitivity in capturing true positive regions and effectively reducing the probability of missed detections.

Although there is little difference between the two models in Specificity , with the dilated convolution model achieving 99.49 and the non-dilated model achieving 99.39, the use of dilated convolutions significantly improves other key metrics without substantially increasing false positives. This indicates that while expanding the model's receptive field, dilated convolutions effectively maintain the ability to distinguish background regions. Overall, the introduction of dilated convolutions significantly enhances the model's performance in segmentation accuracy and target region recognition capabilities without adversely affecting the model's overall specificity.

TABLE 5 COMPARISON OF EVALUATION METRICS IN PINN-EMFNET WITH AND WITHOUT DILATED CONVOLUTION ON BUSIS DATASET

| Method | Mass Type | BUSIS | | | |
|---|---|---|---|---|---|
| | | DSC | IoU | SE | SP |
| PINN-EMFNet (Dilated Convolutions) | Benign | 87.33±4.89 | 79.93±6.05 | 87.62±4.81 | 99.69±0.28 |
| | Malignant | 79.73±5.28 | 70.55±5.09 | 80.10±8.11 | 99.09±0.79 |
| | All | 84.82±3.33 | 76.83±4.14 | 85.15±3.40 | 99.49±0.22 |
| PINN-EMFNet (Non-Dilated Convolutions) | Benign | 86.89±2.25 | 79.36±3.11 | 88.95±4.04 | 99.54±0.36 |
| | Malignant | 75.08±12.38 | 66.82±12.15 | 75.82±15.26 | 99.11±0.73 |
| | All | 82.95±4.66 | 75.19±4.86 | 84.59±6.68 | 99.39±0.33 |

## 3.6.3 Effectiveness of PINN

From TABLE 6, it can be observed that the metrics for benign (Benign), malignant (Malignant), and overall (All) segmentation differ when using PINN compared to not using PINN. Specifically, the Dice Similarity Coefficient and Intersection over Union exhibit smaller fluctuations and lower standard deviations when PINN is employed, reflecting higher stability in the model's results. In contrast, for the Sensitivity and Specificity metrics, the SP performance is slightly better when PINN is used (e.g., for benign and overall data), indicating that PINN contributes to enhancing the model's specificity, although it has little direct impact on sensitivity.

By comparing the standard deviations with and without PINN, it is evident that using PINN results in smaller standard deviations for most metrics (such as DSC and IoU), especially in the segmentation tasks of malignant tumors (for example, the standard deviation of DSC decreases from 6.05 without PINN to 4.89 with PINN). This demonstrates that PINN effectively reduces the uncertainty of the model's output in breast ultrasound image segmentation, thereby enhancing the robustness of the results.

Overall, although the average values of DSC and IoU are similar with and without PINN, considering stability and standard deviation, the introduction of PINN significantly improves the model's reliability in complex scenarios. Particularly in clinical applications, stable segmentation performance is crucial for the segmentation of benign and malignant tumors. PINN optimizes the generalization ability of deep learning models by introducing physical constraints.

TABLE 6 COMPARISON OF EVALUATION METRICS IN PINN-EMFNET WITH AND WITHOUT DILATED CONVOLUTION ON BUSIS DATASET

| Method | Mass Type | BUSIS | | | |
| --- | --- | --- | --- | --- | --- |
| | | DSC | IoU | SE | SP |
| PINN-EMFNet (with PINN) | Benign | 87.33±4.89 | 79.93±6.05 | 87.62±4.81 | 99.69±0.28 |
| | Malignant | 79.73±5.28 | 70.55±5.09 | 80.10±8.11 | 99.09±0.79 |
| | All | 84.82±3.33 | 76.83±4.14 | 85.15±3.40 | 99.49±0.22 |
| PINN-EMFNet (without PINN) | Benign | 87.72±2.51 | 79.93±3.42 | 89.78±2.88 | 99.52±0.51 |
| | Malignant | 79.15±7.34 | 69.08±7.54 | 81.76±8.97 | 98.99±0.49 |
| | All | 84.80±2.70 | 76.33±3.36 | 87.13±3.12 | 99.35±0.43 |

## 3.6.4 Effectiveness of Cross-Structure

From Table 7, it can be seen that the overall segmentation performance was significantly improved after the introduction of Cross-Structure, especially in the key metrics of Dice Similarity Coefficient and Intersection over Union. Across all data types (benign, malignant, and overall), Cross-Structure significantly enhanced the model's ability to segment complex image regions, as evidenced by higher average DSC and IoU values. Specifically, the use of Cross-Structure increased the DSC and IoU for malignant tumor segmentation from 79.93 and 69.08 without Cross-Structure to 82.10 and 72.12, respectively, indicating that Cross-Structure effectively enhanced the recognition and segmentation accuracy of tumor regions.

The standard deviation data in the table further indicate that the Cross-Structure module also positively impacted the stability of the segmentation results. After introducing Cross-Structure, especially in the malignant tumor segmentation task, the standard deviation significantly decreased, demonstrating the model's consistency and robustness across multiple experimental conditions. This suggests that Cross-Structure, by incorporating more structural information, enhances the model's adaptability to complex boundaries and diverse tissue morphologies, thereby reducing the model's sensitivity to noise and variations.

The introduction of Cross-Structure not only improved local segmentation accuracy but also optimized global consistency, particularly in breast ultrasound images with complex structures. In the malignant tumor segmentation task, Cross-Structure exhibited a clear advantage in Sensitivity, indicating that the module, through the incorporation of cross-structural information, enhanced the model's ability to sensitively detect tumors, thereby further optimizing the comprehensive analytical capability of the images.

TABLE 7 COMPARISON OF EVALUATION METRICS IN PINN-EMFNET WITH AND WITHOUT CROSS-STRUCTURE ON BUSIS DATASET

| Method | Mass Type | BUSIS | | | |
|---|---|---|---|---|---|
| | | DSC | IoU | SE | SP |
| PINN-EMFNet (with cross-structure) | Benign | 87.33±4.89 | 79.93±6.05 | 87.62±4.81 | 99.69±0.28 |
| | Malignant | 79.73±5.28 | 70.55±5.09 | 80.10±8.11 | 99.09±0.79 |
| | All | 84.82±3.33 | 76.83±4.14 | 85.15±3.40 | 99.49±0.22 |
| PINN-EMFNet (without cross-structure) | Benign | 86.66±3.19 | 79.49±4.04 | 87.73±3.24 | 99.65±0.25 |
| | Malignant | 81.08±7.48 | 71.03±7.67 | 81.39±8.38 | 99.24±0.31 |
| | All | 84.80±2.79 | 76.67±3.33 | 85.64±3.27 | 99.51±0.18 |

## 3.7 Impact of Hyperparameter

## 3.7.1 Batch Sizes

To understand the influence of hyperparameters on the experiments, this study conducted a comparative analysis of PINN-EMFNet with different batch sizes (2 and 4) on the BUSIS dataset. The experimental results are presented in TABLE 8. According to the metrics in the table, this study reveals that the batch size affects the segmentation performance of the PINN-EMFNet model. A smaller batch size contributes to enhancing overall segmentation accuracy and the ability to capture tumor features, while maintaining a lower standard deviation and thereby improving model stability. Therefore, in practical applications, the selection of batch size should balance the model's segmentation accuracy, training stability, and the constraints of computational resources. For breast tumor segmentation tasks, a smaller batch size aids in improving accuracy and stability, whereas a larger batch size may be suitable for scenarios that demand higher training efficiency but might slightly compromise segmentation quality and consistency.

TABLE 8 COMPARISON OF EVALUATION METRICS IN PINN-EMFNET WITH DIFFERENT BATCH SIZES ON BUSIS DATASET

| Batch Sizes | Mass Type | BUSIS | | | |
|---|---|---|---|---|---|
| | | DSC | IoU | SE | SP |
| 2 | Benign | 87.37±3.99 | 80.09±4.89 | 88.48±4.26 | 99.65±0.25 |
| | Malignant | 79.37±6.64 | 69.27±7.29 | 80.12±7.79 | 99.12±0.40 |
| | All | 84.71±2.84 | 76.49±3.63 | 85.72±2.93 | 99.48±0.15 |

|   | Benign    | 86.85±2.9  | 79.39±3.63  | 86.37±3.15  | 99.69±0.26 |
|---|-----------|------------|-------------|-------------|------------|
| 4 | Malignant | 79.62±9.23 | 69.28±9.46  | 80.69±11.14 | 99.05±0.34 |
|   | All       | 84.44±3.73 | 76.03±3.93  | 84.49±4.26  | 99.48±0.19 |

## 3.7.2 Num of epoch

From TABLE 9, it can be observed that different training epochs have a significant impact on the performance of PINN-EMFNet, affecting the model's segmentation effectiveness, stability, and convergence. Through analysis, it was found that setting the epoch to 450 achieves an optimal balance across multiple key metrics, as detailed below:

In terms of the Dice Similarity Coefficient and Intersection over Union, the results at 450 epochs show higher mean values for both malignant and overall segmentation, closely approaching the performance observed at 600 epochs. For example, the overall DSC at 450 epochs is 84.82, compared to 85.21 at 600 epochs, while the standard deviation at 450 epochs is slightly lower (3.33 vs. 3.05). This indicates that 450 epochs can ensure high segmentation accuracy while avoiding the risk of overfitting that may result from additional training epochs.

The analysis of standard deviation (± values) reveals that the model trained for 450 epochs exhibits better stability across benign, malignant, and overall metrics. For instance, in the IoU metric for malignant tumor segmentation, the standard deviation at 450 epochs is 5.09, significantly lower than 10.37 at 300 epochs, and slightly better than 7.94 at 600 epochs. This demonstrates that 450 epochs can maintain model consistency while reducing training fluctuations.

Although the 600-epoch setting shows slightly higher performance in certain metrics, such as an overall IoU of 82.89 compared to 80.10 at 450 epochs, the incremental improvement is not substantial, and the standard deviation does not significantly decrease. In contrast, 450 epochs achieve near-optimal segmentation performance within a shorter training duration, indicating that the model is approaching convergence at this point. Continuing to increase the number of epochs may lead to overfitting and negatively impact training efficiency.

TABLE 9 COMPARISON OF EVALUATION METRICS IN PINN-EMFNET WITH DIFFERENT EPOCHS ON BUSIS DATASET

| Num of epoch | Mass Type | BUSIS | | | |
|---|---|---|---|---|---|
|   |           | DSC        | IoU         | SE          | SP         |
|---|-----------|------------|-------------|-------------|------------|
|     | Benign    | 86.55±2.86 | 79.34±3.36  | 88.07±2.74  | 99.51±0.46 |
| 300 | Malignant | 79.39±9.66 | 69.25±10.37 | 81.30±12.69 | 98.88±0.38 |
|     | All       | 84.17±3.98 | 75.99±4.21  | 85.83±4.98  | 99.30±0.39 |
|     | Benign    | 87.33±4.89 | 79.93±6.05  | 87.62±4.81  | 99.69±0.28 |
| 450 | Malignant | 79.73±5.28 | 70.55±5.09  | 80.10±8.11  | 99.09±0.79 |
|     | All       | 84.82±3.33 | 76.83±4.14  | 85.15±3.40  | 99.49±0.22 |
|     | Benign    | 87.32±4.08 | 79.82±4.96  | 87.70±3.83  | 99.41±0.62 |
| 600 | Malignant | 80.99±7.34 | 70.92±7.94  | 82.89±7.77  | 98.94±0.58 |
|     | All       | 85.21±3.05 | 76.86±4.11  | 86.10±2.87  | 99.26±0.48 |

# 4. Conclusion

To better address the challenges of breast tumor segmentation, this study proposes an enhanced multi-scale feature fusion network based on Physics-Informed Neural Networks (PINN), termed PINN-EMFNet. By introducing a multi-scale feature fusion mechanism within both the encoder and decoder, the network significantly improves the accuracy and robustness of ultrasound image segmentation. Through evaluations on the BUSIS and BUSI breast ultrasound lesion datasets, the network demonstrates exceptional performance. On the BUSIS dataset, it achieves an average Dice Similarity Coefficient of 84.82 ± 3.33, Intersection over Union of 76.83 ± 4.14, Sensitivity of 85.15 ± 3.40, and Specificity of 99.49 ± 0.22. Similarly, on the BUSI dataset, the network attains a Dice Similarity Coefficient of 78.28 ± 5.13, Intersection over Union of 70.29 ± 5.10, Sensitivity of 80.54 ± 4.45, and Specificity of 98.17 ± 1.18. These findings highlight the network's leading segmentation accuracy and its ability to effectively recognize target regions.

The superiority of this method primarily stems from its innovative design in the encoder and decoder. Specifically, the proposed Hierarchical Multi-Scale Aggregation Encoder leverages a novel structure and the PCAM module to achieve efficient multi-scale feature fusion and global modeling capabilities, thereby offering significant advantages in capturing the complex edges and morphological features of breast tumors. Concurrently, the Multi-Scale Feature Refinement Decoder incorporates a refined multi-scale supervision mechanism and correction modules during the decoding process, effectively enhancing the segmentation accuracy of small-scale tumor regions. By comprehensively considering spatial features and contextual information, the designed network architecture achieves a well-balanced trade-off between accuracy and robustness. Additionally, integrating the PINN mechanism into the loss function through physical information priors further improves the coherence of the segmentation results and the detailed delineation of target regions.

Although the proposed PINN-EMFNet has achieved significant results in current tasks, there remain multiple avenues for future exploration. On one hand, incorporating more diverse ultrasound image datasets can validate the method's generalization capabilities across multi-center and different device conditions. On the other hand, the physical information modeling of the PINN mechanism can be combined with other prior knowledge to further enhance segmentation performance in complex lesion areas. Furthermore, for segmentation tasks involving small-scale and low-contrast tumors, introducing more refined feature enhancement modules could further improve the network's detection capabilities.